\documentclass[runningheads]{llncs}

\usepackage{eccv}

\usepackage{eccvabbrv}

\usepackage{graphicx}
\usepackage{xcolor}
\usepackage{booktabs}
\usepackage{multirow}
\usepackage{makecell}
\usepackage{algorithm}
\usepackage{algpseudocode}
\usepackage{arydshln}
\usepackage[accsupp]{axessibility}  

\usepackage{subcaption}
\providecommand{\Lcdnv}{\mathcal{L}_{\mathrm{CDNV}}}
\providecommand{\Lord}{\mathcal{L}_{\mathrm{ord}}}

\usepackage{hyperref}

\usepackage{orcidlink}

\begin{document}

\title{Ordinal Neural Collapse as a Representation Prior for Visual Navigation} 

\titlerunning{ORION}

\author{E-In Son$^{\star}$\orcidlink{0009-0003-9668-8355} \and
Jung-Taak Kim$^{\star}$\orcidlink{0009-0003-0975-8598} \and
Seung-Woo Seo$^{\dagger}$\orcidlink{0000-0003-4890-8563}}

\authorrunning{E.~Son et al.}

\institute{Dept. of ECE \& INMC,
Seoul National University\\
\email{\{pingpang,mychoco333,seo\}@snu.ac.kr}}



\maketitle

\let\oldthefootnote\thefootnote
\let\thefootnote\relax
\footnotetext{$^{\star}$\,Equal contribution.\\$^{\dagger}$\,Corresponding author.}\let\thefootnote\oldthefootnote

\begin{abstract}
\label{sec:abstract}

Learning robust navigation policies directly from visual observations remains a fundamental challenge in vision-based robotic navigation. In end-to-end imitation learning approaches, the visual encoder and action decoder are jointly optimized using a single action loss, which provides only an indirect supervisory signal to the encoder. This indirect supervision frequently results in the encoder learning ambiguous, action-agnostic representations. The problem is further complicated by substantial variations in scene structure and appearance across diverse environments, as well as the prevalence of visual distractors inherent to real-world navigation settings. Such action-agnostic features cause the navigation policy to produce inconsistent actions at ambiguous decision points, leading to navigation failure.
To overcome these limitations, we propose ORION (Ordinal Neural Collapse for Visual Navigation), a method that explicitly organizes the encoder's representation space according to the ordinal structure of navigation actions.
In the context of goal-directed navigation, ego-centric control categories from \textit{Far Left} to \textit{Far Right} exhibit a natural ordinal relationship in which neighboring classes share similar visual contexts, while semantically opposing classes differ substantially in appearance. 
We encourage class representations to be arranged sequentially along a single discriminative axis, while suppressing off-axis variance within each class.
The pretrained encoder is then integrated into a diffusion-based navigation framework, and the full pipeline is fine-tuned end-to-end.
Extensive experiments in both simulation and real-world settings show that ORION consistently outperforms end-to-end and neural collapse baselines in navigation success rate and goal progress, with notable gains in visually challenging scenarios such as complex multi-way intersections.

\end{abstract}
\section{Introduction}
\label{sec:intro}


Recently, end-to-end imitation learning has become a dominant approach for vision-based navigation. By jointly optimizing a vision encoder and an action decoder from expert demonstrations, recent systems have achieved strong performance across diverse environments~\cite {hirose2019deep, Shah2023ViNTAF, chi2023diffusionpolicy}. Diffusion-based action decoders have further improved performance by capturing multimodal action distributions~\cite{chi2023diffusionpolicy, sridhar2024nomad}.

\begin{figure}[t!]
    \centering
    \begin{subfigure}[b]{0.42\textwidth}
        \includegraphics[width=\textwidth]{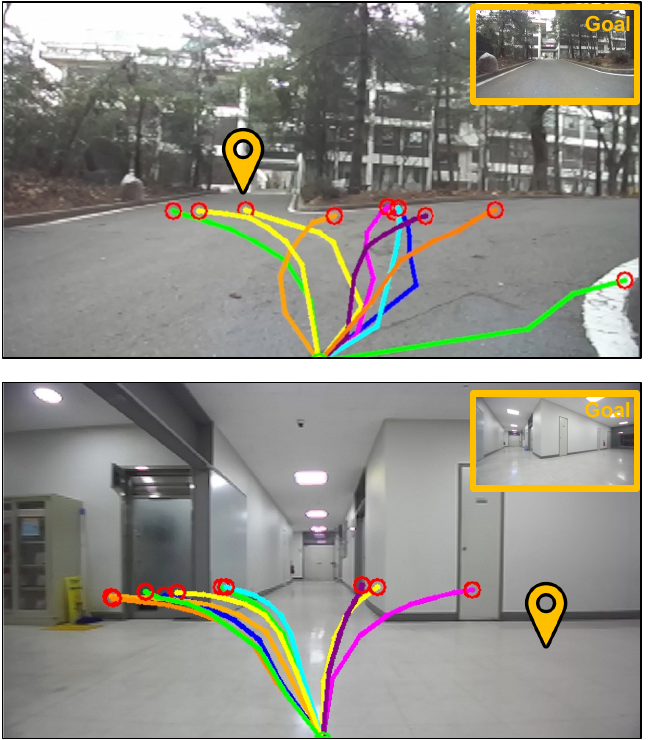}
        \caption{NoMaD (Baseline)}
        \label{fig1:nomad}
    \end{subfigure}%
    \hspace{4pt}
    \begin{subfigure}[b]{0.42\textwidth}
        \includegraphics[width=\textwidth]{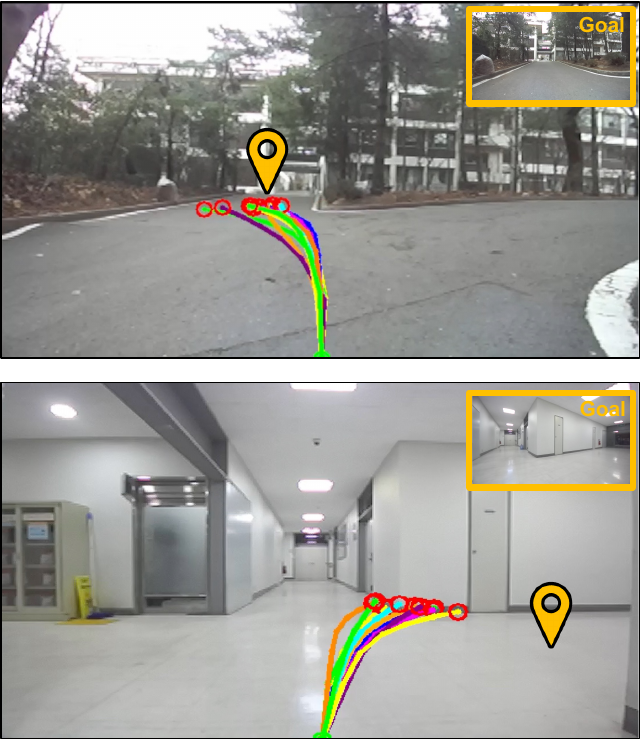}
        \caption{ORION (Ours)}
        \label{fig1:ours}
    \end{subfigure}
    \caption{\textbf{Structured representations improve action consistency at intersections. } Given the same goal image, (a)~the NoMaD baseline generates candidate waypoints scattered across multiple directions, while (b)~our method concentrates them toward the goal direction. Each colored trajectory represents one action candidate sampled from the action decoder.}
    \label{fig:fig1_teaser}
\end{figure}

However, relying solely on a single action loss provides limited learning signals for the vision encoder. Since the objective is to imitate actions, the training loss acts on the final action output rather than supervising the intermediate representation. Prior work has shown that encoders trained this way often latch onto visually salient but task-irrelevant features~\cite{zhang2021learning, stooke2021decoupling}.
In visual navigation, this problem is worse because training data spans diverse environments with wide variation in layout, lighting, and terrain~\cite{mirowski2017learning, codevilla2018end}.
The encoder must compress this visual diversity into a shared latent space, yet receives no explicit signal to organize its representations around control-relevant structure.
As a result, the learned representations lack consistent action-related semantics, leading to degraded performance in novel visual conditions.
Fig.~\ref{fig1:nomad} illustrates this effect. At an intersection, the baseline policy produces candidate paths scattered across multiple directions rather than consistently toward the goal. This suggests that the model focuses on imitating expert actions, while the learned representation may not sufficiently capture goal-relevant visual structure.

Recent work has explored structuring visual representations to be \emph{control-aware}, decoupling representation learning from policy learning through staged training~\cite{laskin2020curl, stooke2021decoupling, zhang2021learning, nair2023r3m, majumdar2023where}.
A particularly compelling perspective comes from \emph{neural collapse} (NC)~\cite{papyan2020prevalence}, which characterizes the terminal phase of training in deep classifiers: features of the same class converge to their class mean, and the class means form a symmetric simplex geometry.
Notably, Qi~\etal~\cite{qi2025controloriented} observed that neural collapse also emerges in vision-based control tasks formulated with discrete control classes, and showed that explicitly encouraging this structure through encoder pretraining improves downstream policy performance.
Their work raises a natural question: what is the appropriate class definition and geometric structure for a visual navigation task?


In navigation, the robot's immediate control action is largely determined by where it needs to go. This intuition motivates the use of goal-relative directions as a natural basis for defining discrete control classes, which can be divided into ordered categories ranging from \textit{Far Left} to \textit{Far Right}~\cite{pomerleau1988alvinn}.
Unlike categorical control classes where all pairs are treated as equally spaced, goal-relative classes are naturally more similar to neighboring classes than to distant ones.
Neighboring classes correspond to nearby steering commands and share similar visual contexts, while opposing classes represent different navigational situations.
We leverage this natural ordering as a prior structure for the learned representation space. Recent theoretical analysis has shown that ordinal classes naturally collapse onto a one-dimensional ordered subspace, rather than the simplex geometry assumed in standard neural collapse~\cite{ma2025neural}.


Building on this, we propose \textbf{ORION} (Ordinal Neural Collapse for Visual Navigation), which explicitly imposes ordinal geometric structure on a vision encoder's representation space for continuous visual control. To our knowledge, this represents the first attempt to leverage ordinal neural collapse geometry for visual navigation.


Rather than restricting the encoder's capacity, our method provides a structured initialization from which downstream end-to-end training can refine the full range of navigation-relevant features.
Specifically, we develop a supervised ordinal projection that aligns class means along a learned axis, paired with a compactness objective that suppresses variance perpendicular to the axis.
We build on the NoMaD~\cite{sridhar2024nomad} navigation framework, which combines a vision encoder with a Transformer context module, and a diffusion action decoder~\cite{chi2023diffusionpolicy}.
Our approach operates in two stages: we first pretrain the vision encoder using ordinal neural collapse objectives on direction-labeled navigation data, then fine-tune the entire pipeline end-to-end with the standard diffusion policy loss.


In simulation and real-world navigation experiments, ORION demonstrates improved robustness compared to the end-to-end trained baseline, particularly in novel environments and visually ambiguous scenarios such as intersections with similar visual features. \\


Our contributions are as follows:
\begin{itemize}
    \item We identify that goal-relative navigation classes have inherently non-uniform inter-class similarity, forming a natural ordinal structure. Existing control-oriented methods overlook this property by assuming uniform class separation.
    \item We propose ORION, a pretraining method for ordinal control classes, comprising supervised ordinal projection and orthogonal variance suppression, that enforces ordinal alignment while collapsing representations toward the ordinal axis.
    \item We evaluate our method in simulation and real-world navigation experiments and show that ORION improves the success rate by up to +26 pp and reduces heading jerk by up to 71\% over the NoMaD baseline.
\end{itemize}
\section{Related Work}
\label{sec:related}


\subsection{Representation Learning for Robot Control}
\label{sec:related_repr}

While end-to-end imitation learning jointly optimizes perception and control, it often yields encoders that overfit to task-irrelevant visual distractors rather than actionable structures~\cite{zhang2021learning, geirhos2020shortcut, ilyas2019adversarial}.
This has motivated staged training approaches that decouple representation learning from policy optimization through contrastive objectives~\cite{stooke2021decoupling, laskin2020curl}, task-aware pretraining on large-scale data~\cite{nair2023r3m}, or systematic benchmarking of pretrained visual backbones for embodied tasks~\cite{majumdar2023where}.
These works establish that control-relevant representations improve downstream performance, but they do not prescribe a specific target geometry for the representation space.


\subsection{Neural Collapse and Control-Oriented Clustering}
\label{sec:related_nc}

\subsubsection{Neural Collapse in Classification.}
Papyan~\etal~\cite{papyan2020prevalence} discovered that deep classifiers in their terminal training phase exhibit a striking regularity: within-class features collapse to their class mean, and the class means converge to a Simplex Equiangular Tight Frame (ETF) with uniform pairwise cosine similarity $\cos\theta = -1/(K{-}1)$.
This geometry has been theoretically grounded~\cite{zhu2021geometric, zhou2022optimization} and extended to imbalanced settings~\cite{fang2021exploring, hong2024neural}.
Critically, NC has evolved from an observation into a prescriptive tool, where explicitly enforcing NC geometry improves robustness and generalization in downstream tasks~\cite{yang2022inducing, li2024understanding}.

\subsubsection{Control-Oriented Clustering.}
Qi~\etal~\cite{qi2025controloriented} brought this prescriptive perspective to visual control.
They observed that NC-like clustering spontaneously arises in behavior cloning when features are labeled by action-derived classes, and proposed Relative Pose Orthants (REPOs)---a multi-dimensional binary partition yielding $2^d$ categorical classes---as control-oriented labels.
Pretraining encoders to minimize NC metrics improved downstream policy performance by 10--35\% across multiple tasks.
Because REPO classes are categorical with no ordinal relationship, the Simplex ETF geometry is structurally appropriate: all class pairs \emph{should} be equidistant.

\subsubsection{Ordinal Neural Collapse.}
Ma~\etal~\cite{ma2025neural} proved that in ordinal regression, class means collapse onto a one-dimensional subspace ordered by class rank, rather than forming a Simplex ETF.
Their analysis relies on cumulative probabilities and threshold parameters specific to ordinal regression; we adopt the geometric insight---ordinal alignment on a 1D subspace---as an inductive bias, not the theoretical framework itself (Section~\ref{sec:method}).

\begin{figure}[t]
    \centering
    \includegraphics[width=0.8\linewidth]{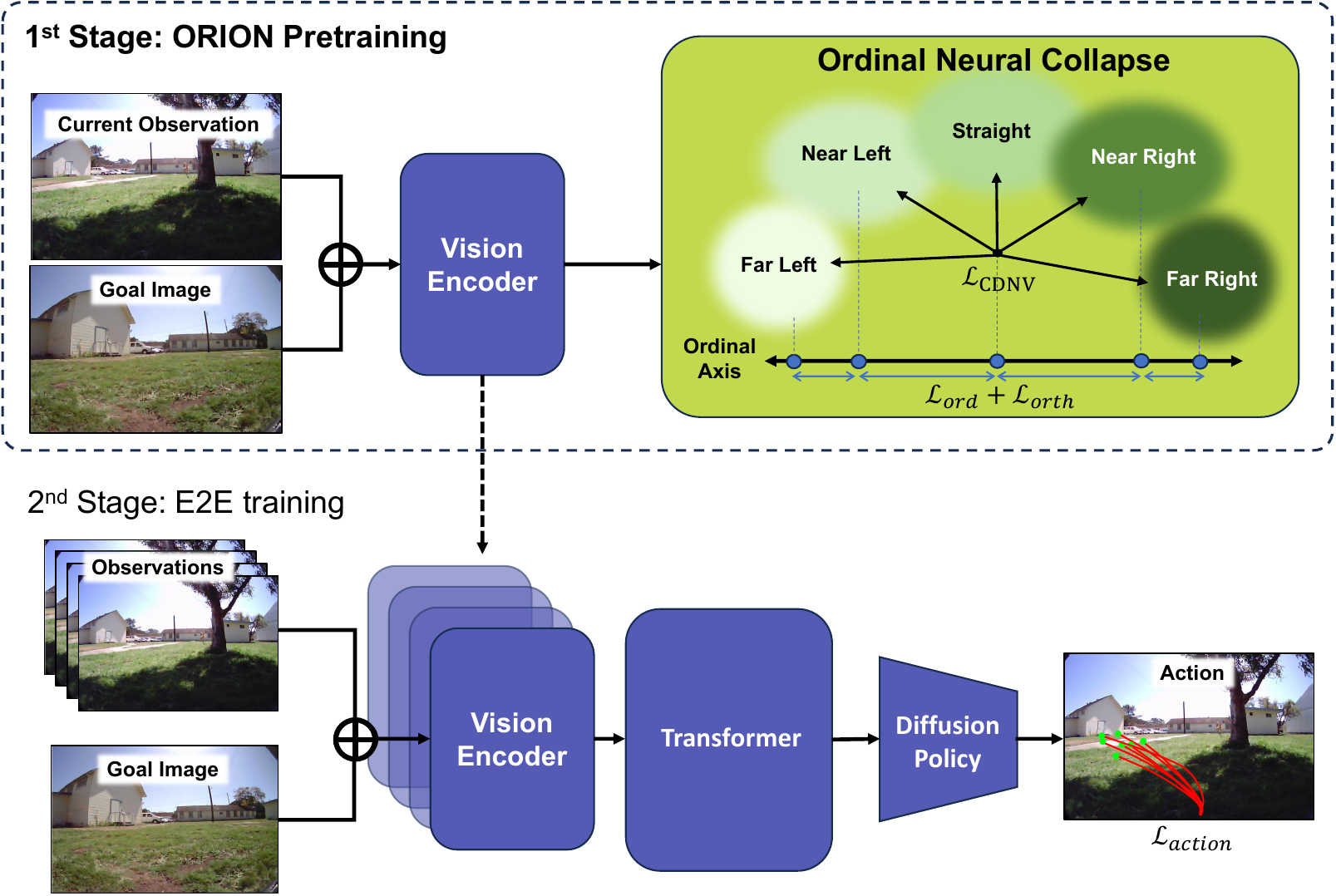}
    \caption{
    \textbf{Overall pipeline of ORION.}
    In Stage~1, ORION pretraining aligns the encoder feature space with the ordinal relationships among navigation actions. In Stage~2, the pretrained encoder is plugged into the NoMaD pipeline and fine-tuned end-to-end with the diffusion policy loss.}
    \label{fig:fig2_overall_pipeline}
    \vspace{-0.25cm}
\end{figure}


\subsection{Learning-Based Visual Navigation}
\label{sec:related_nav}

Learning-based navigation has progressed from classical modular pipelines~\cite{thrun2005probabilistic} and early visuomotor policies~\cite{pomerleau1988alvinn, mirowski2017learning, codevilla2018end} toward scalable, data-driven approaches.
ViNT~\cite{Shah2023ViNTAF} trained a Transformer-based model on diverse multi-robot data for topological goal-reaching. NoMaD~\cite{sridhar2024nomad} extended this with a diffusion policy decoder~\cite{chi2023diffusionpolicy} that represents multimodal action distributions, unifying goal-conditioned navigation and exploration through goal masking.
NoMaD architecture serves as our baseline.
Subsequent work has further refined the diffusion-based action decoder through cost-guided sampling~\cite{zeng2025navidiffusor} and bridge models with informative priors~\cite{ren2025navibridger}.

Despite advances in policy architecture and sampling strategy, the encoder in these systems is trained end-to-end without explicit structure in the representation space.
Hong~\etal~\cite{hong2023learning} addressed this partially through semantic map supervision for indoor navigation, but no prior work has organized encoder representations to reflect the ordinal structure of the navigation action space.
Our approach fills this gap by pretraining the vision encoder with ordinal alignment objectives before policy learning.
\section{Method}
\label{sec:method}

\subsection{Problem Setup and Overview}
\label{sec:overview}

We consider image-goal navigation via imitation learning: given an observation $o_t$ and a goal image $g$, the agent predicts a continuous action $a_t \in \mathbb{R}^d$.
We build on the NoMaD architecture~\cite{sridhar2024nomad}, which decomposes the policy into a vision encoder $f_\theta$ that maps each observation and the goal to a per-frame feature $h_t = f_\theta(o_t, g) \in \mathbb{R}^D$, a Transformer context module $T_\phi$ that temporally aggregates the per-frame features into $z_t = T_\phi(h_{t-p:t})$, and a diffusion decoder $D_\psi$~\cite{chi2023diffusionpolicy} that generates actions conditioned on $z_t$.

When such an architecture is trained end-to-end, the encoder is supervised only indirectly through the action denoising loss. 
Our approach adds a pretraining stage before this end-to-end pipeline.
In Stage~1, we train only the encoder $f_\theta$ using the Ordinal Neural Collapse (ONC) objectives, which shape the geometry of the feature space to better align with the structure of visual navigation tasks.
In Stage~2, the pretrained encoder is integrated into the full NoMaD pipeline and fine-tuned end-to-end with the diffusion policy loss.
We emphasize that pretraining provides a structured geometric initialization; it does not constrain what the encoder learns during downstream training.
The overall pipeline is illustrated in Fig.~\ref{fig:fig2_overall_pipeline}.

\subsection{Goal-Relative Control Classes}
\label{sec:control_class}

To define supervisory structure for representation shaping, we first discretize the continuous action space into ordered classes based on the angular direction toward the goal.
Let $\alpha_t \in [-\pi, \pi)$ denote the relative angular offset between the agent's heading and the goal direction.
We partition the front field of view into $K$ uniform angular bins and assign each sample to a class $y_t \in \{0, \dots, K-1\}$; if the goal lies outside the field of view, the sample is assigned to a separate out-of-view class $y_t = K$, yielding $K+1$ classes in total.

These classes have a natural ordinal structure: class $0$ corresponds to the far right goal direction, class $K-1$ to the far left, and intermediate classes are ordered between them.
Consequently, neighboring classes share similar steering contexts, while distant classes represent different behaviors.
The out-of-view class is excluded from the ordinal objectives, which assume an ordered relationship between classes. It contributes only to $\mathcal{L}_{\mathrm{CDNV}}$, which does not rely on any ordinal assumption.

\subsection{ORION Pretraining Objectives}
\label{sec:onc_objectives}

We propose ORION pretraining loss, which consists of three components: class separability, supervised ordinal projection, and orthogonal compactness.
Let $\mu_k$ denote the mean feature of class $k$, $\bar{\mu}$ the global mean, $\tilde{\mu}_k := \mu_k - \bar{\mu}$ the centered class mean, and $\sigma_k^2$ the within-class variance.

\subsubsection{Class Separability.}
\label{sec:cdnv}
To encourage compact class clusters, we minimize the class-distance normalized variance (CDNV)~\cite{galanti2021role}, which measures intra-class compactness relative to inter-class separation:
\begin{equation}
    \label{eq:CDNV}
    \mathcal{L}_{\mathrm{CDNV}}
    =
    \frac{1}{|\mathcal{P}|}
    \sum_{(k,k') \in \mathcal{P}}
    \frac{
    \sigma_k^2 + \sigma_{k'}^2
    }{
    2 \left\| \tilde{\mu}_k - \tilde{\mu}_{k'} \right\|_2^2
    },
\end{equation}
where $\mathcal{P} = \{(k,k') \mid k < k'\}$ is the set of all unordered class pairs.

\subsubsection{Supervised Ordinal Projection.}
\label{sec:supervised_proj}

Our method aligns class means along a one-dimensional axis according to their ordinal order.
This design is motivated by ONC theory~\cite{ma2025neural}, which shows that under ordinal regression, class means collapse onto a one-dimensional subspace ordered by class rank.

We assign each class an equally spaced target position $t_k = k - (K-1)/2$ for $k \in \{0, \dots, K-1\}$, and define the supervised ordinal axis as the direction that best aligns the centered class means with the target ordering:

\begin{equation}
    \label{eq:axis}
    w = \frac{\tilde{M}^\top \mathbf{t}}{\|\tilde{M}^\top \mathbf{t}\|_2},
\end{equation}

where $\tilde{M} = [\tilde{\mu}_0, \dots, \tilde{\mu}_{K-1}]^\top \in \mathbb{R}^{K \times D}$ is the matrix of centered class means and $\mathbf{t} = [t_0, \dots, t_{K-1}]^\top$.
The ordinal projection loss then encourages the projected class means $p_k = \tilde{\mu}_k^\top w$ to match the targets:

\begin{equation}
    \label{eq:ord_loss}
    \mathcal{L}_{\mathrm{ord}}
    =
    \frac{1}{K}\sum_{k=0}^{K-1}\left( p_k - t_k \right)^2.
\end{equation}

Unlike PCA-based approaches that identify the principal axis without label information, our supervised projection directly incorporates the known ordinal structure, guaranteeing zero ordinal violations by construction.

\subsubsection{Orthogonal Compactness.}
\label{sec:orth_var}
To concentrate features onto the ordinal subspace, we penalize within-class variance in the directions orthogonal to the ordinal axis:
\begin{equation}
    \label{eq:orth_loss}
    \mathcal{L}_{\mathrm{orth}}
    =
    \frac{1}{N}\sum_{n=1}^{N}\left\| (I - ww^\top)(h_n - \mu_{y_n}) \right\|^2,
\end{equation}
where $w$ is the supervised ordinal axis defined in Eq.~\eqref{eq:axis}, and $N$ is the number of (non-OOV) samples in the mini-batch.
While $\mathcal{L}_{\mathrm{ord}}$ positions class means along $w$, $\mathcal{L}_{\mathrm{orth}}$ collapses the orthogonal directions, encouraging features to lie on the ordinal axis.
This geometric constraint applies only during pretraining. Downstream end-to-end fine-tuning is free to recover orthogonal variation as needed for the full navigation task.

\subsubsection{Overall Objective.}
The full pretraining loss is:
\begin{equation}
    \label{eq:total_loss}
    \mathcal{L}_{\mathrm{ORION}}
    =
    \lambda_{\mathrm{cdnv}}\, \mathcal{L}_{\mathrm{CDNV}}
    +
    \lambda_{\mathrm{ord}}\, \mathcal{L}_{\mathrm{ord}}
    +
    \lambda_{\mathrm{orth}}\, \mathcal{L}_{\mathrm{orth}},
\end{equation}
where $\mathcal{L}_{\mathrm{CDNV}}$ promotes class separability, $\mathcal{L}_{\mathrm{ord}}$ enforces ordinal alignment along a learned axis, and $\mathcal{L}_{\mathrm{orth}}$ reduces orthogonal variation to reinforce one-dimensional structure.
The full training procedure is shown in Alg.~\ref{alg:onc}.

\subsection{Training Considerations}
\label{sec:training}

{
\begin{algorithm}[t]
\caption{ORION Pretraining and Downstream Fine-tuning}
\label{alg:onc}
\begin{algorithmic}[1]
\Require Dataset $\mathcal{D} = \{(o_t, g, a_t, y_t)\}$, encoder $f_\theta$, context module $T_\phi$, diffusion decoder $D_\psi$

\Statex \textbf{Stage 1: ORION Pretraining} (train $f_\theta$ only)
\State Initialize running class means $\{\mu_k\}$, variances $\{\sigma_k^2\}$, axis $w$
\For{each mini-batch $\mathcal{B}$}
    \State $h_t \gets f_\theta(o_t,g)$ for all $(o_t, y_t) \in \mathcal{B}$
    \State Update $\{\mu_k\}$, $\{\sigma_k^2\}$ via Welford's algorithm
    \State Compute centered class means $\tilde{\mu}_k \gets \mu_k - \bar{\mu}$
    \State Compute $\mathcal{L}_{\mathrm{CDNV}}$ \Comment{Eq.~\eqref{eq:CDNV}}
    \State Compute supervised axis: $w \gets \mathrm{EMA}\!\left(w,\; \tilde{M}^\top \mathbf{t} / \|\tilde{M}^\top \mathbf{t}\|\right)$
    \State Compute $\mathcal{L}_{\mathrm{ord}}$ from projections $p_k = \tilde{\mu}_k^\top w$ \Comment{Eq.~\eqref{eq:ord_loss}}
    \State Compute $\mathcal{L}_{\mathrm{orth}} = \frac{1}{N}\sum_{n=1}^{N}\left\| (I - ww^\top)(h_n - \mu_{y_n}) \right\|^2$ \Comment{Eq.~\eqref{eq:orth_loss}}
    \State Update $\theta$ by $\nabla_\theta \mathcal{L}_{\mathrm{ORION}}$ \Comment{Eq.~\eqref{eq:total_loss}}
\EndFor

\Statex
\Statex \textbf{Stage 2:} Fine-tune $f_\theta, T_\phi, D_\psi$ end-to-end with diffusion policy loss~\cite{sridhar2024nomad}.
\end{algorithmic}
\end{algorithm}
}

\subsubsection{Encoder-level Application.}
We apply ORION pretraining to the encoder features $h_t$ rather than the context-level features $z_t$.
The context module integrates temporal and goal information through attention. Applying geometric constraints at this stage disrupts temporal aggregation and results in unstable feature norms. Additional experimental results are included in the supplementary material.

\subsubsection{Stable Statistics Estimation.}
The pretraining objectives depend on class means and variances, which can be noisy when estimated from single mini-batches.
We maintain running statistics using Welford's online algorithm~\cite{welford1962note} and smooth the ordinal axis $w$ via exponential moving average across batches.
Gradient flow is preserved by computing losses from current-batch features centered with the incrementally updated statistics (details in supplementary material).
\section{Experiments}
\label{sec:exp}

\subsection{Experimental Setup}
\label{sec:exp_setup}

\subsubsection{Training Data and Control Classes.}

We use the same training dataset as NoMaD~\cite{sridhar2024nomad}, which consists of trajectories collected from diverse environments and robotic platforms.
The dataset includes RECON~\cite{shah2021rapid}, SCAND~\cite{karnan2022socially}, GoStanford~\cite{hirose2019deep}, and SACSoN~\cite{hirose2023sacson}, totaling approximately 140 hours of navigation time. 
Goal images are sampled from future observations along the same trajectory, drawn 3–20 steps ahead of the current observation during training.
Each sample is labeled with a goal-relative class based on the angular offset $\alpha_t$ between the agent's heading and the goal.
We partition the front field of view $[-90^\circ, 90^\circ]$ into $K=5$ uniform angular bins and assign samples outside this range to an out-of-view class, yielding six classes in total. This explicit out-of-view class captures goal-irrelevant states, allowing the encoder to convey such conditions to downstream control.

\subsubsection{Training.}

In Stage~1, we pretrain the EfficientNet-B0~\cite{tan2019efficientnet} encoder for 30 epochs using AdamW~\cite{loshchilov2018decoupled} with a learning rate of $10^{-5}$, cosine scheduling, and batch size 256.
Loss weights are $\lambda_{\mathrm{cdnv}} = 1.0$, $\lambda_{\mathrm{ord}} = 1.0$, $\lambda_{\mathrm{orth}} = 1.0$.
In Stage~2, we integrate the pretrained encoder into the full NoMaD pipeline and fine-tune all parameters end-to-end for 30 epochs with the diffusion policy loss, following the NoMaD defaults~\cite{sridhar2024nomad}: a 1D conditional U-Net decoder with 10 denoising steps, trajectory prediction horizon of 8, and learning rate $10^{-4}$.

\subsubsection{Baselines.}

We compare our method with the following baselines:
\begin{itemize}
    \item \textbf{ViNT}~\cite{Shah2023ViNTAF}: A goal-conditioned policy trained end-to-end using a single action regression loss.
    \item \textbf{NoMaD}~\cite{sridhar2024nomad}: The diffusion-based policy trained end-to-end without encoder pretraining.
    \item \textbf{NC-ETF}: Encoder pretraining based on the Simplex ETF geometry proposed by Qi~\etal~\cite{qi2025controloriented}. 
    We adapt this formulation to the visual navigation setting, enforcing three standard NC metrics (CDNV, STD~Norm, STD~Angle).
\end{itemize}

\subsubsection{Evaluation Metrics.}

We report success rate, progress, collision, and heading jerk.
\begin{itemize}
    \item \textbf{Success Rate (SR):} the fraction of trials in which the agent reaches the goal. The goal-reaching distance tolerance is set to $1.0$\,m for indoor, and $5.0$\,m for outdoor. The time limit is set to $300$\,s per episode.
    \item \textbf{Progress (Prog.):} the ratio of visited visual subgoal nodes to the total number of subgoal nodes along the route, capturing partial task completion in failed episodes.
    \item \textbf{Collision:} the mean number of physical contacts with obstacles per episode, indicating navigation safety.
    \item \textbf{Heading Jerk (H-Jerk):} the median absolute change in heading angle between consecutive time steps. This metric captures short-horizon directional stability and is particularly diagnostic at intersections, where temporally inconsistent encoder features manifest as rapid heading oscillations.
\end{itemize}

{
\begin{table*}[t]
\centering
\caption{\textbf{Navigation performance in simulated environments.}
We evaluate in Indoor (2D-3D-S) and Outdoor (Gazebo Citysim) environments. \textbf{Bold} denotes the best result per route.}
\label{tab:table1_sim}
\setlength{\tabcolsep}{4pt}
\renewcommand{\arraystretch}{1.15}
\resizebox{\textwidth}{!}{%
\begin{tabular}{cclcccc}
\toprule
\multirow{2}{*}{\textbf{Environment}} & \multirow{2}{*}{\textbf{Route}} & \multirow{2}{*}{\textbf{Method}}
  & \multicolumn{3}{c}{\textit{Task Performance}} & \textit{Traj. Quality} \\
\cmidrule(lr){4-6} \cmidrule(lr){7-7}
  & &
  & \textbf{SR} (\%) $\uparrow$
  & \textbf{Prog.} (\%) $\uparrow$
  & \textbf{Collision} $\downarrow$
  & \textbf{H-Jerk (med.)} $\downarrow$ \\
\midrule
\multirow{8}{*}{\shortstack{Indoor\\(2D-3D-S)}}
  & \multirow{4}{*}{Basic}
  & ViNT~\cite{Shah2023ViNTAF}             & 76  & 74.58  & 0.24  & 4.12 \\
  & & NoMaD~\cite{sridhar2024nomad}        & 92  & 95.17  & 0.02  & 4.61 \\
  & & NC-ETF                               & 86  & 89.32  & 0.14  & 4.33 \\
  & & ORION                  & \textbf{96}  & \textbf{96.68}  & \textbf{0.04}  & \textbf{1.58} \\
\cmidrule(lr){2-7}
  & \multirow{4}{*}{Intersection}
  & ViNT~\cite{Shah2023ViNTAF}             & 0  & 36.12  & 1.00  & 4.99 \\
  & & NoMaD~\cite{sridhar2024nomad}        & 17  & 32.07  & 0.81  & 15.48 \\
  & & NC-ETF                               & 12  & 25.48  & 0.88  & 12.13 \\
  & & ORION                  & \textbf{91}  & \textbf{93.82}  & \textbf{0.09}  & \textbf{4.99} \\
\midrule
\multirow{8}{*}{\shortstack{Outdoor\\(Citysim)}}
  & \multirow{4}{*}{Basic}
  & ViNT~\cite{Shah2023ViNTAF}             & 45  & 73.58  & 1.48  & 1.22 \\
  & & NoMaD~\cite{sridhar2024nomad}        & 88  & 89.85  & 0.12  & 5.08 \\
  & & NC-ETF                               & 89  & 91.18  & 0.21  & 2.38 \\
  & & ORION                  & \textbf{94}  & \textbf{94.68}  & \textbf{0.06}  & \textbf{1.09} \\
\cmidrule(lr){2-7}
  & \multirow{4}{*}{\shortstack{Multi\\Intersection}}
  & ViNT~\cite{Shah2023ViNTAF}             & 16  & 44.24  & 2.16  & \textbf{2.34} \\
  & & NoMaD~\cite{sridhar2024nomad}        & 48  & 70.30  & 1.80  & 34.49 \\
  & & NC-ETF                               & 40  & 68.57  & 1.39  & 11.98 \\
  & & ORION                  & \textbf{74}  & \textbf{86.39}  & \textbf{0.45}  & 10.13 \\
\bottomrule
\end{tabular}%
}
\end{table*}
}

\subsection{Simulation Results}
\label{sec:sim_results}

We evaluate our method in two simulated environments: (i) Stanford 2D-3D-S~\cite{armeni2017joint}, an indoor environment with narrow corridors, and (ii) a Gazebo-based urban simulation (Citysim)~\cite{koenig2004design}. Each environment includes two test routes with $N=100$ trials per route. All methods are deployed on a Clearpath Husky UGV.

\subsubsection{Overall Performance.}

Across both simulation environments, ORION consistently outperforms the baselines in terms of success rate and progress.
(Tab.~\ref{tab:table1_sim})
On \textit{Basic}, which is mostly straight with a single intersection, all diffusion-based methods achieve high success rates, with ORION showing moderate improvement over NoMaD (e.g., 94 vs.\ 88 SR on Citysim \textit{Basic}). 
Performance differences become more pronounced on \textit{Multi-Intersection} spanning approximately 200\,m and requiring four consecutive directional decisions, where ORION outperforms NoMaD in success rate by a margin of +26 pp (48\% to 74\%).
ViNT performs worse than both diffusion-based methods across settings.

\subsubsection{Intersection-Level Behavior.}

\begin{figure}[t]
    \centering
    \includegraphics[width=\linewidth]{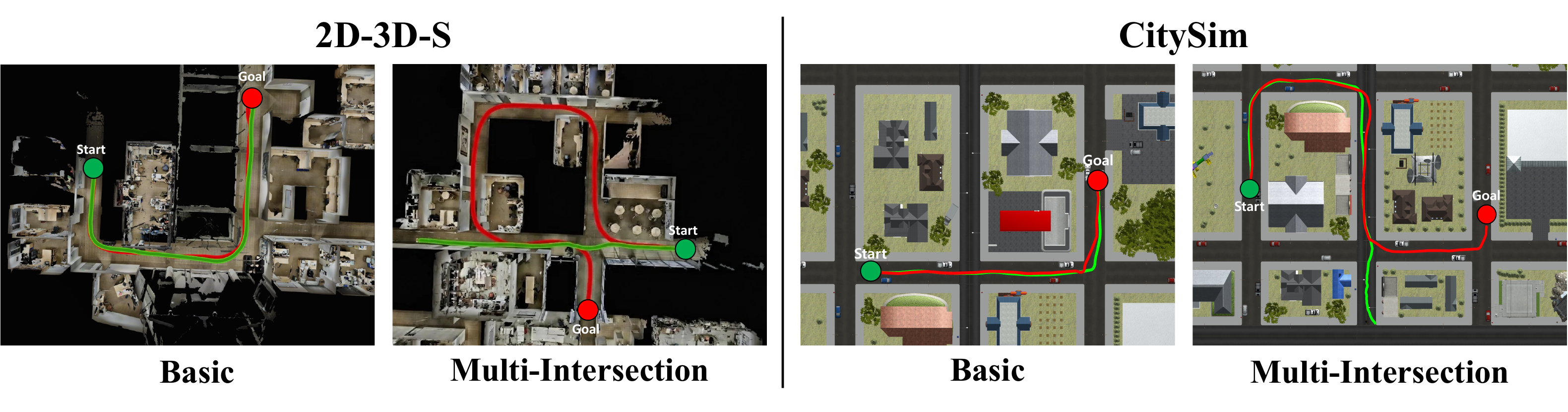}
    \caption{\textbf{Qualitative comparison in simulation.}
    Green and red dots denote the start and goal positions, respectively. Trajectories are color-coded as NoMaD (green) and ORION (red).}
    \label{fig:fig3_sim}
\end{figure}

Fig.~\ref{fig:fig3_sim} compares top-view trajectories of NoMaD and ORION at an intersection. 
NoMaD exhibits temporal flickering: the diffusion decoder alternates between competing directions across successive time steps. ORION instead concentrates on candidate paths toward a single direction, producing more consistent trajectories and reducing heading jerk by 71\% over NoMaD (34.49 → 10.13 on \textit{Multi-Intersection}).

\subsubsection{Geometric Interpretation.}

NC-ETF, which enforces Simplex ETF geometry during pretraining, performs comparably to or below the NoMaD baseline despite using the same two-stage training procedure as ORION.
This indicates that representation pretraining with a mismatched geometric structure does not improve navigation and can even degrade performance.

\subsubsection{Failure Analysis.}

ORION failures on intersection routes fall into two categories: (i) premature turning, where the agent initiates a turn too early and collides with obstacles at the intersection boundary; and (ii) branch misjudgment, where the agent selects an incorrect direction.
Premature turning is also observed in NoMaD and mainly arises from imprecise turn timing rather than representation quality. Branch misjudgment is the dominant failure mode of NoMaD; ORION reduces its frequency but does not eliminate it.

\subsection{Real-World Results}
\label{sec:real_results}

We evaluate our method in three complex real-world environments: \textit{Indoor, Campus, and Field.}
We conduct 10 evaluation episodes per route on \textit{Indoor} and \textit{Campus}. 
On \textit{Field} environment, due to longer travel distances and reduced controllability of environmental factors, we conduct 6 trials per route.
All policies are deployed on a Clearpath Husky UGV with a ZED 2i stereo camera (left RGB stream only) and an NVIDIA Jetson Orin for onboard inference. 


\begin{table*}[]
\centering
\caption{\textbf{Navigation performance in real-world environments.} We evaluate in Indoor, Campus, and Field environments. \textbf{Bold} denotes the best result per route.}
\label{tab:table2_real}

\begin{tabular*}{\textwidth}{
@{\extracolsep{\fill}}
l
ccc
ccc
ccc
@{}
}
\toprule
& \multicolumn{3}{c}{Indoor}
& \multicolumn{3}{c}{Campus}
& \multicolumn{3}{c}{Field} \\
\cmidrule(lr){2-4}
\cmidrule(lr){5-7}
\cmidrule(lr){8-10}

Method
& SR$\uparrow$ & Prog.$\uparrow$ & H-Jerk$\downarrow$
& SR$\uparrow$ & Prog.$\uparrow$ & H-Jerk$\downarrow$
& SR$\uparrow$ & Prog.$\uparrow$ & H-Jerk$\downarrow$ \\
\midrule

ViNT
& 20 & 56.97 & \textbf{2.34}
& 40 & 52.15 & \textbf{7.76}
& 0 & 36.70 & \textbf{8.51} \\

NoMaD
& 40 & 57.19 & 5.09
& 60 & 73.95 & 13.67
& 16.67 & 37.41 & 20.77 \\

ORION
& \textbf{60} & \textbf{92.12} & 4.64
& \textbf{70} & \textbf{80.61} & 12.17
& \textbf{33.33} & \textbf{48.99} & 13.45 \\

\bottomrule
\end{tabular*}
\end{table*}

\subsubsection{Overall Performance.}

Across all environments, our method achieves higher success rates and progress compared to the baselines in real-world experiments (Tab.~\ref{tab:table2_real}).

\textit{Indoor} is particularly challenging due to visually repetitive corridors where all choices appear nearly identical. Here, ORION achieves a higher success rate than both baselines, better distinguishing similar visual features.

On \textit{Campus}, the same intersection instability observed in the simulation also appears. Both ViNT and NoMaD often make inconsistent directional decisions at visually ambiguous junctions.
In contrast, ORION exhibits more stable behavior at intersections, leading to more reliable navigation outcomes.

\textit{Field} environment involves long-range navigation ($\approx$200m) over uneven terrain with sparse visual features.
In this setting, all methods show relatively low success rates. Nevertheless, ORION consistently achieves higher progress toward the goal, indicating more stable long-horizon navigation without drifting away from the intended direction.

\subsubsection{Qualitative Analysis.}

\begin{figure*}[t]
    \centering
        \centering
        \includegraphics[width=\linewidth]{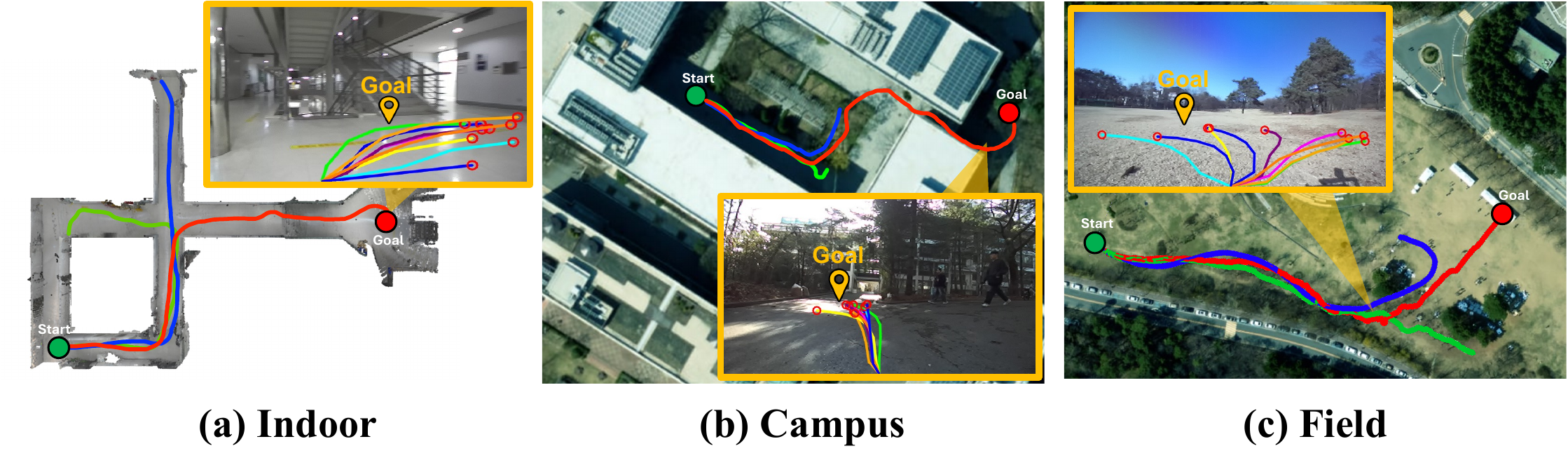}
    \caption{
    \textbf{Real-world qualitative results.}
    Green and red dots denote start and goal positions, respectively. Trajectories are color-coded as ViNT (blue), NoMaD (green), and ORION (red). The orange frames show example ego images and their corresponding sampled paths in each trajectory.
    }
    \label{fig:fig4_real}
\end{figure*}

Fig.~\ref{fig:fig4_real} shows qualitative comparisons across the three environments.

On \textit{Indoor}, our policy frequently attempts to avoid the staircase obstacle, occasionally failing to reach the goal as a result. This obstacle avoidance behavior, which is not explicitly supervised during pretraining, emerges from the end-to-end fine-tuning stage. The robot generally follows the subgoal sequence and achieves high progress, with failures caused by physical constraints rather than navigation errors.

On \textit{Campus}, baseline policies show temporal flickering near intersections, consistent with simulation observations. ORION trajectories show more consistent path selection at these junctions (See first row of Fig.~\ref{fig:fig1_teaser}).

\textit{Field} presents a large open space where many feasible directions exist, making it easy for the robot to drift from the intended route.
ORION is able to recover and return toward the route even after temporary deviations, a behavior not observed with the baselines.

Notably, ViNT achieves the lowest H-Jerk across all environments. However, this mainly reflects the limitation of its regression-based policy. By averaging over multimodal action distributions, the policy avoids decisive turns. As a result, ViNT produces smooth but less responsive trajectories, which leads to the lowest success rates.
In contrast, ORION achieves lower H-Jerk than NoMaD despite using the same diffusion-based decoder. This suggests that the pretrained encoder helps the policy make clearer goal-relevant decisions, leading to more decisive actions. 
Additional qualitative results are provided in the supplementary material.

\subsection{Ablation Studies}
\label{sec:ablation}

We ablate three key design choices underlying ORION: its representation geometry, the individual loss terms, and the two-stage training structure.

\subsubsection{Geometry vs. Direction Labels.}

To isolate whether ORION's gain stems from its ordinal-collapse geometry or merely from adding a direction-aware pretraining stage, we compare against CE-only pretraining: cross-entropy classification over the same $K{=}5$ directional bins, followed by the identical Stage-2 protocol. As shown in
Tab.~\ref{tab:abl-geom}, CE-only reaches only 16\% SR on \textit{Multi-Intersection},
below the no-pretraining NoMaD baseline (48\%), and yields the highest
heading jerk. Direction supervision here is not merely
insufficient but actively harmful, as forcing directional labels through a
mismatched classification objective degrades the representation rather than
structuring it. ORION's gains therefore require its task-matched
ordinal-collapse objectives, not direction supervision alone.

\begin{table}[h]
\centering
\small
\setlength{\tabcolsep}{4pt}
\caption{\textbf{Pretraining design ablations.} We evaluate on Citysim \textit{Multi-Intersection}, reporting a mean over 100 episodes.}
\label{tab:table3_ablation}
\vspace{-0.5cm}
\begin{subtable}[t]{0.43\linewidth}
  \centering
  \caption{Geometry vs.\ direction labels.}
  \label{tab:abl-geom}
  \begin{tabular}{lccc}
  \toprule
  Method & SR$\uparrow$ & Prog.$\uparrow$ & H-Jerk$\downarrow$ \\
  \midrule
  NoMaD          & 48 & 70.30 & 34.49 \\
  CE-only        & 16 & 38.35 & 54.63 \\
  ORION          & 74 & 86.39 & 10.13 \\
  \bottomrule
  \end{tabular}
\end{subtable}
\hfill
\begin{subtable}[t]{0.55\linewidth}
  \centering
  \caption{Loss component contributions.}
  \label{tab:abl-loss}
  \begin{tabular}{lccc}
  \toprule
  Variant & SR$\uparrow$ & Prog.$\uparrow$ & H-Jerk$\downarrow$ \\
  \midrule
  $\Lcdnv$ only            & 33 & 55.54 & 19.41 \\
  $\Lcdnv + \Lord$         & 45 & 69.16 & 42.98 \\
  ORION (full)             & 74 & 86.39 & 10.13 \\
  \bottomrule
  \end{tabular}
\end{subtable}
\end{table}

\subsubsection{Loss Components.}

Tab.~\ref{tab:abl-loss} reports ablation results with individual loss terms removed.
With $\mathcal{L}_{\mathrm{CDNV}}$ alone, the model achieves only 33\% SR. Adding $\mathcal{L}_{\mathrm{ord}}$ improves SR to 45\% by ordering class means along a 1D axis. The full model reaches 74\% SR, as $\mathcal{L}_{\mathrm{orth}}$ tightens within-class distributions around the axis that $\mathcal{L}_{\mathrm{ord}}$ defines. 
$\mathcal{L}_{\mathrm{orth}}$ requires $\mathcal{L}_{\mathrm{ord}}$ 
to be present, as it reduces orthogonal variance relative to the axis that 
$\mathcal{L}_{\mathrm{ord}}$ constructs.

\subsubsection{Two-Stage vs.\ Joint Training.}

Tab.~\ref{tab:table4_joint} compares two-stage ORION pretraining against NC-Joint, which applies ORION losses as auxiliary objectives during end-to-end training from the start.
NC-Joint fails entirely, achieving 0\% success rate with H-Jerk an order of magnitude higher than NoMaD (147.95 vs.\ 5.08).
Qualitatively, the NC-Joint policy generates paths that spread radially regardless of goal direction, failing to concentrate toward the target.
This indicates that the ORION and diffusion objectives interfere destructively when optimized simultaneously: neither the encoder's ordinal structure nor the decoder's action prediction converges properly under conflicting gradients.
Two-stage pretraining resolves this by separating the two objectives temporally, allowing the encoder to establish geometric structure in Stage~1 before the diffusion objective is introduced in Stage~2. (See supplementary material for further details)

\begin{table}[]
\centering
\caption{
    \textbf{Two-stage vs.\ joint training.} We evaluate on the Citysim \textit{Basic}, 
    reporting a mean over 100 episodes.
}
\label{tab:table4_joint}
\small
\begin{tabular}{l ccc}
\toprule
\textbf{Variant} & SR $\uparrow$ & Prog. $\uparrow$ & H-Jerk $\downarrow$ \\
\midrule
NoMaD & 88 & 89.85 & 5.08 \\
NC-Joint & 0 & 10.34 & 147.95  \\
ORION (Two-Stage) & 94 & 94.68 & 1.09 \\
\bottomrule
\end{tabular}
\end{table}

\subsection{Representation Analysis}
\label{sec:repr_analysis}

\begin{figure}[t]
    \centering
    \includegraphics[width=\linewidth]{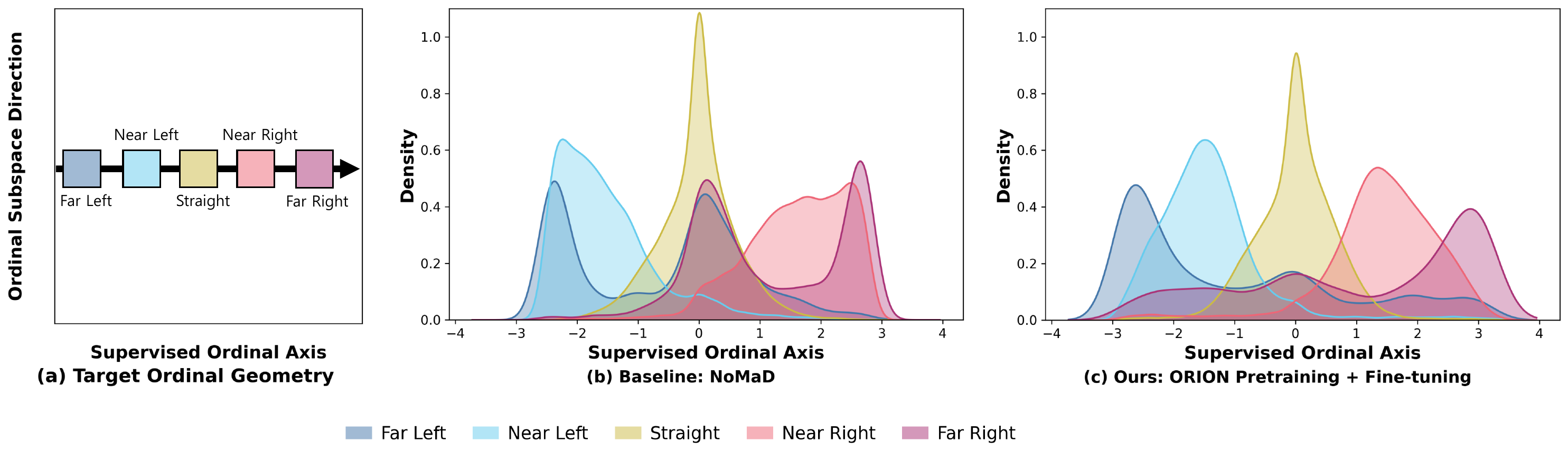}
    \caption{
        \textbf{Class distributions on the supervised ordinal axis after Stage~2 fine-tuning.}
        (a)~Target ordinal geometry.
        (b)~NoMaD: Far and Near classes nearly overlap at both extremes.
        (c)~ORION: all five classes maintain distinct modes.
        Projections are normalized to zero mean and unit variance. Distributions are computed on the held-out RECON\cite{shah2021rapid} test split using Gaussian kernel density estimation (bandwidth $h = 0.15$). Residual overlap near boundaries reflects bin-boundary ambiguity, where visually similar samples across an angular bin boundary receive different labels.
    }
    \label{fig:fig5_representation}
\end{figure}

Section~\ref{sec:ablation} shows that joint training fails to establish ordinal structure.
A natural question is whether Stage~2 fine-tuning preserves the structure established during 
pretraining.
Fig.~\ref{fig:fig5_representation} addresses this question by comparing class distributions projected onto the supervised ordinal axis after full training (Stage~1 + Stage~2).
Both models achieve correct ordinal ordering by mode, indicating that end-to-end training itself induces partial ordinal structure.
However, the degree of within-class concentration differs substantially.
In NoMaD (Fig. 5b), class distributions are broad with heavy cross-class overlap near the center of the axis.
In ORION (Fig. 5c), each class occupies a more compact region with Far--Near mode separations $6{-}9\times$ larger than in NoMaD (0.37--0.53 vs.\ $<0.06$ in normalized units).
This confirms that the ordinal structure from Stage~1 persists through 30 epochs of end-to-end fine-tuning, validating the two-stage design.

\section{Conclusion}
\label{sec:conclusion}

We present ORION (Ordinal Neural Collapse for Visual Navigation) pretraining framework that embeds ordinal geometric structure into the vision encoder's representation space for image-goal navigation.
ORION pretraining shapes the encoder feature space to reflect the ordinal similarity of steering decisions, providing a structured initialization for the downstream diffusion policy.
Experiments in both simulation and real-world deployments show that ORION consistently improves navigation success rates over end-to-end and NC baselines, with particular gains at visually ambiguous intersections.
Our results highlight the importance of structured representations for interpreting visual observations in end-to-end navigation models. We believe this principle can extend to other control tasks where actions exhibit inherent structure.


\section*{Acknowledgements}
This work was supported by the Challengeable Future Defense Technology Research and Development Program through the Agency for Defense Development (ADD) funded by the Defense Acquisition Program Administration (DAPA) (No. 915108201), and the BK21 FOUR Program of the Education and Research Program for Future ICT Pioneers, Seoul National University in 2026.

%
%
\bibliographystyle{splncs04}
\bibliography{refs}
\end{document}


\title{Ordinal Neural Collapse as a Representation Prior for Visual Navigation\\[1mm]
{\large\mdseries Supplementary Material}}

\titlerunning{ORION}

\author{E-In Son$^{\star}$\orcidlink{0009-0003-9668-8355} \and
Jung-Taak Kim$^{\star}$\orcidlink{0009-0003-0975-8598} \and
Seung-Woo Seo$^{\dagger}$\orcidlink{0000-0003-4890-8563}}

\authorrunning{E.~Son et al.}

\institute{Dept. of ECE \& INMC, Seoul National University\\
\email{\{pingpang,mychoco333,seo\}@snu.ac.kr}}

\maketitle

\section{Method Design Justification}
\label{sec:method_just}

\subsection{Effect of the Number of Ordinal Bins ($K$)}
\label{sec:K_bins}

We empirically evaluate how the number of ordinal bins affects navigation 
performance by varying $K \in \{3, 5, 7\}$ while keeping all other settings 
identical. Results are summarized in Tab.~\ref{tab:bin_result}.

$K{=}5$ achieves the best performance across both routes. $K{=}3$ partitions 
the front FOV into 60° bins, too coarse to separate visually distinct steering 
decisions, and falls substantially below the NoMaD baseline. $K{=}7$ narrows 
each bin to approximately 26°, at which point angular estimation noise 
frequently crosses class boundaries and degrades label quality. $K{=}5$ 
balances directional resolution against label stability.

{
\setlength{\tabcolsep}{6pt}
\begin{table}[h]
    \centering
    \caption{\textbf{Effect of the number of ordinal bins ($K$).} 
    Navigation on CitySim \textit{Basic} and \textit{Multi-Intersection} (100 episodes each).}
    \label{tab:bin_result}
    \begin{tabular}{ccccccc}
    \toprule
    \multirow{2}{*}{\textbf{$K$}} & \multicolumn{3}{c}{Basic} & 
    \multicolumn{3}{c}{Multi-Intersection} \\
    \cmidrule(lr){2-4} \cmidrule(lr){5-7}
     & SR $\uparrow$ & Prog. $\uparrow$ & Collision $\downarrow$ 
     & SR $\uparrow$ & Prog. $\uparrow$ & Collision $\downarrow$ \\
    \midrule
    NoMaD & 88 & 89.85 & 0.12 & 48 & 70.30 & 1.80 \\
    \midrule
    3                    & 36 & 52.99 & 1.80 & 2  & 17.63 & 3.18 \\
    5 (ours) & \textbf{94} & \textbf{94.68} & \textbf{0.06} & \textbf{74} & \textbf{86.39} & \textbf{0.45} \\
    7                    & 83 & 87.20 & 0.64 & 50 & 69.38 & 1.05 \\
    \bottomrule
    \end{tabular}
    \vspace{-0.5cm}
\end{table}}

\subsection{Bin Discretization: Equal-Angle vs.\ K-Means Clustering}
\label{sec:kmeans}

\begin{figure}[t]
    \centering
    \includegraphics[width=\textwidth]{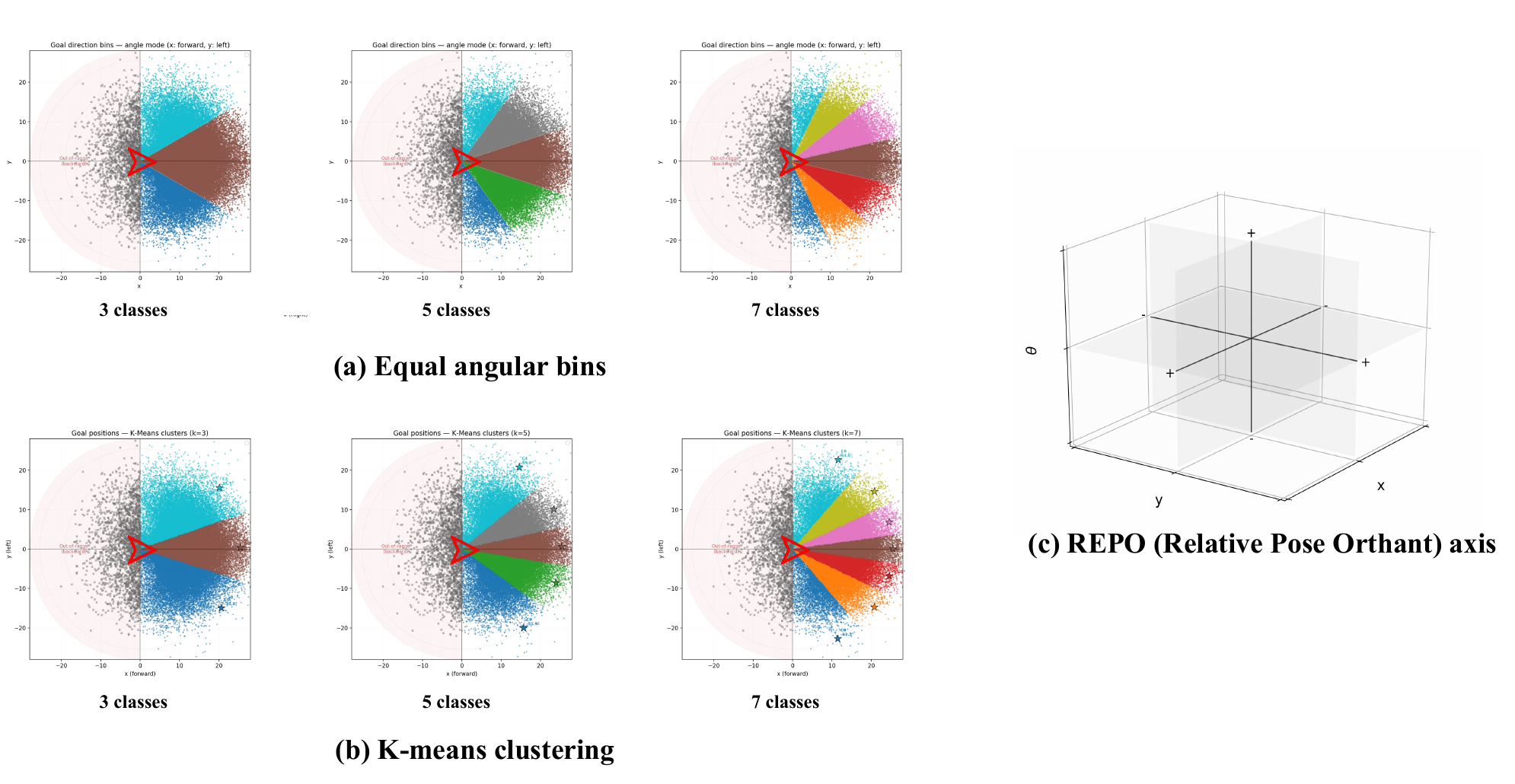}
    \caption{\textbf{Comparison of discretization methods for goal-relative directions.}\newline (a) equal angular bins, (b) k-means clustering, and (c) illustration of REPO axis.}
    \label{fig:goal_bins}
\end{figure}

The main paper adopts equal angular bins as a parameter-free definition that directly reflects the angular structure of navigation directions.
To examine whether a data-driven alternative yields meaningfully different boundaries, we derive bin boundaries from k-means clustering of goal-relative angles in the training data. 
Fig.~\ref{fig:goal_bins}(a,b) shows that the k-means clustering-based partitions closely match the equal-angle definition.
We therefore use the equal-angle definition in all main experiments due to its simplicity.

\subsection{REPO Discretization}
\label{sec:repo}

\begin{figure}[h]
    \centering
    \includegraphics[width=\textwidth]{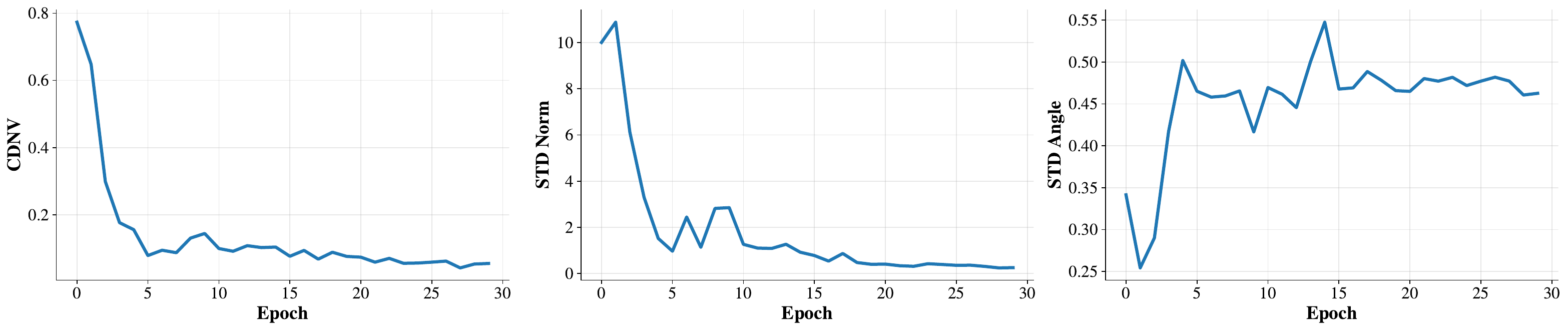}
    \caption{\textbf{Training curves of REPO-ETF.} CDNV and STDNorm decrease during training, but STDAngle fails to converge, preventing the formation of the ETF structure.}
    \label{fig:repo_pretrain}
\end{figure}

We additionally evaluate the REPO (Relative Pose Orthant) discretization 
proposed in prior work~\cite{qi2025controloriented}. REPO defines action 
classes based on the sign of the relative pose components $(x,y,\theta)$, 
partitioning the 3D relative pose space into eight orthants 
(Fig.~\ref{fig:goal_bins}(c)). We adapt this discretization to the navigation 
setting to examine whether ETF geometry remains a viable pretraining target 
when the class structure is not ordinal. As discussed in the main paper, 
REPO orthants do not reflect the angular continuity of navigation directions.
Applying the three standard Neural Collapse losses (CDNV, STDNorm, STDAngle) 
results in unstable optimization in which the STDAngle term fails to converge, 
preventing the feature space from forming a regular simplex (See Fig.~\ref{fig:repo_pretrain}). This further 
supports the main paper's argument that ETF geometry is structurally 
mismatched to tasks with inherently non-uniform inter-class similarity.

\subsection{NC Application Site: Encoder vs.\ Transformer}
\label{sec:enc_vs_ctx}

In the NoMaD architecture, the vision encoder $f_\theta$ produces per-image tokens, which are aggregated by a Transformer context module $T_\phi$ into a context vector $z_t$.
A natural question is whether to apply ONC objectives to the \emph{encoder output} (our choice) or the \emph{transformer output}.

We compare three configurations:
\begin{enumerate}[label=(\alph*)]
    \item \textbf{Encoder NC (ours):} ONC losses applied to the goal-image encoder output $h_g = f_\theta(o_t, g)$.
    \item \textbf{Context NC:} ONC losses applied to the context vector $z_t = T_\phi(h_{t-p:t})$.
    \item \textbf{Context NC + LayerNorm (LN):} Same as (b) with LayerNorm normalization to prevent norm explosion.
\end{enumerate}

Experimental results are shown in Tab.~\ref{tab:enc_vs_ctx}.
Context NC suffers from severe norm explosion (110 vs.\ 25), confirming that the Transformer's attention dynamics conflict with geometric constraints on feature norms. Adding LayerNorm stabilizes the norm but still yields poor navigation performance (SR 27), suggesting that the interference is not merely a scaling issue. Forcing ordinal structure onto context features appears to disrupt the temporal aggregation that the Transformer is designed to perform.

\begin{table}[h]
    \centering
    \caption{\textbf{NC application site comparison.} Navigation results on CitySim \textit{Multi-Intersection} (100 episodes).}
    \label{tab:enc_vs_ctx}
    \begin{tabular}{lc@{\hspace{10pt}} c@{\hspace{10pt}}c c c c}
    \toprule
    \multirow{2}{*}{NC Site} & \multicolumn{2}{c}{Pretrain (Stage~1)} & \multicolumn{3}{c}{Navigation (Stage~2)} \\
    \cmidrule(lr){2-3} \cmidrule(lr){4-6}
     & Feat. Norm & CDNV & SR $\uparrow$ & Prog. $\uparrow$ & Collision $\downarrow$ \\
    \midrule
    Encoder NC (ours) & 25.13 & 0.00269 & \textbf{74} & \textbf{86.39} & \textbf{0.45} \\
    Context NC & 110.52 & 0.00157 & 14 & 24.07 & 2.86 \\
    Context NC + LN & 12.23 & 0.00160 & 30 & 64.55 & 1.80 \\
    \bottomrule
    \end{tabular}
    \vspace{-0.3cm}
\end{table}

\subsection{Joint Training Failure Analysis}
\label{sec:nc_joint}

\begin{figure}[H]
    \centering
    \includegraphics[width=\textwidth]{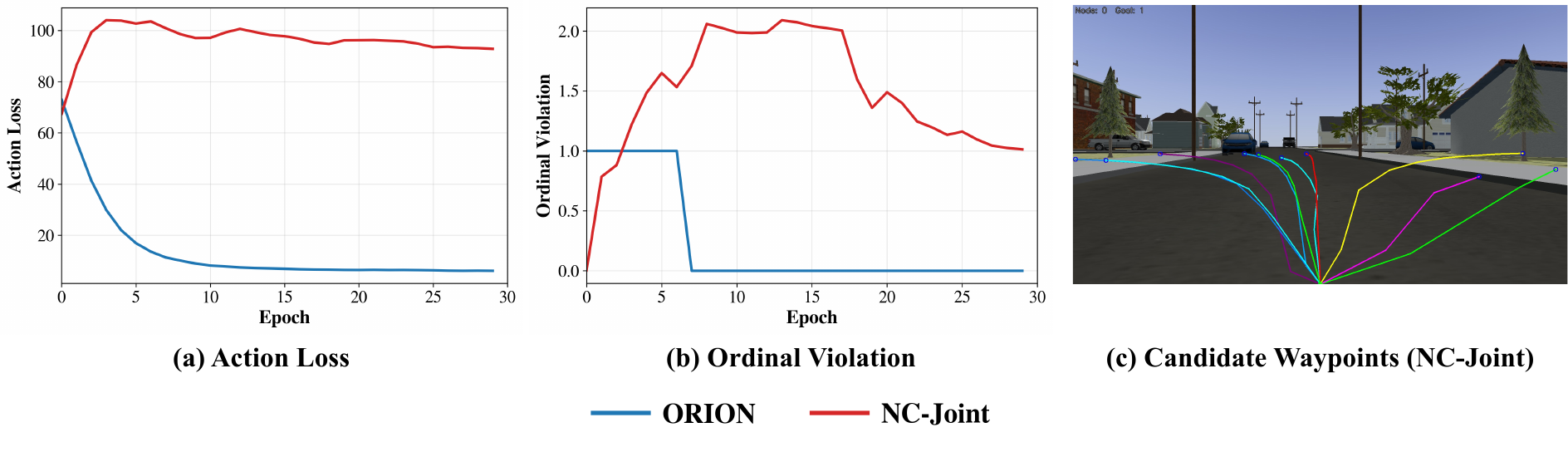}
    \caption{\textbf{Joint training failure.} (a)~Action loss: NC-Joint (red) diverges while ORION (blue) converges. (b)~NC-Joint shows persistently high ordinal violation throughout training, whereas ORION converges to zero. (c)~Resulting candidate waypoints spread radially, ignoring goal direction.}
    \label{fig:joint}
\end{figure}

Tab. 4 in the main paper reports that NC-Joint achieves 0\% success rate.
Fig.~\ref{fig:joint} reveals the underlying cause. When NC and diffusion objectives are optimized simultaneously on a shared encoder, neither converges. 
In particular, the action loss fails to converge (Fig.~\ref{fig:joint} (a)), indicating that the policy is unable to learn a stable imitation policy.
Consequently, the policy produces radial trajectories that do not correspond to valid navigation behavior, leading to a 0\% success rate.

\subsection{Loss Component Ablation (Extended)}
\label{sec:loss_ablation_ext}

Tab. 3 of the main paper reports three ablation variants. 
Here, we introduce two additional variants to further isolate the contribution of each component  ($\mathcal{L}_\mathrm{ord} + \mathcal{L}_\mathrm{orth}$ and $\mathcal{L}_\mathrm{CDNV} + \mathcal{L}_\mathrm{orth}$).

Removing $\mathcal{L}_\mathrm{CDNV}$ leads to the largest performance drop (SR $74{\to}19$), 
indicating that ordinal alignment alone cannot produce a meaningful structure without inter-class separation. 
The variant $\mathcal{L}_\mathrm{CDNV} + \mathcal{L}_\mathrm{orth}$ (SR 35) performs comparably to 
$\mathcal{L}_\mathrm{CDNV}$ alone (SR 33), confirming that $\mathcal{L}_\mathrm{orth}$ provides little 
benefit without the ordinal axis established by $\mathcal{L}_\mathrm{ord}$. 
All incomplete variants fall below the NoMaD baseline (SR 48), highlighting the necessity of the full loss combination.

\begin{table}[h]
    \centering
    \setlength{\tabcolsep}{5pt}
    \caption{\textbf{Extended loss ablation}. Results on CitySim \textit{Multi-Intersection} (100 episodes each).}
    \label{tab:loss_ext}
    \begin{tabular}{lccc}
    \toprule
    Variant & SR $\uparrow$ & Prog. $\uparrow$ & H-Jerk $\downarrow$ \\
    \midrule
    ORION (full)                                         & \textbf{74} & \textbf{86.39} & \textbf{10.13} \\
    $\mathcal{L}_\mathrm{ord} + \mathcal{L}_\mathrm{orth}$  & 19 & 35.68 & 14.83 \\
    $\mathcal{L}_\mathrm{CDNV} + \mathcal{L}_\mathrm{orth}$ & 35 & 60.13 & 12.27 \\
    $\mathcal{L}_\mathrm{CDNV} + \mathcal{L}_\mathrm{ord}$  & 45 & 69.16 & 42.98 \\
    $\mathcal{L}_\mathrm{CDNV}$ only                        & 33 & 55.54 & 19.41 \\
    \bottomrule
    \end{tabular}
    \vspace{-0.8cm}
\end{table}
\section{Additional Analysis of the Learned Representation}
\label{sec:analysis}

We provide additional analysis to understand the structure of the learned representation better. 

\begin{figure}[h]
    \centering
    \includegraphics[width=\linewidth]{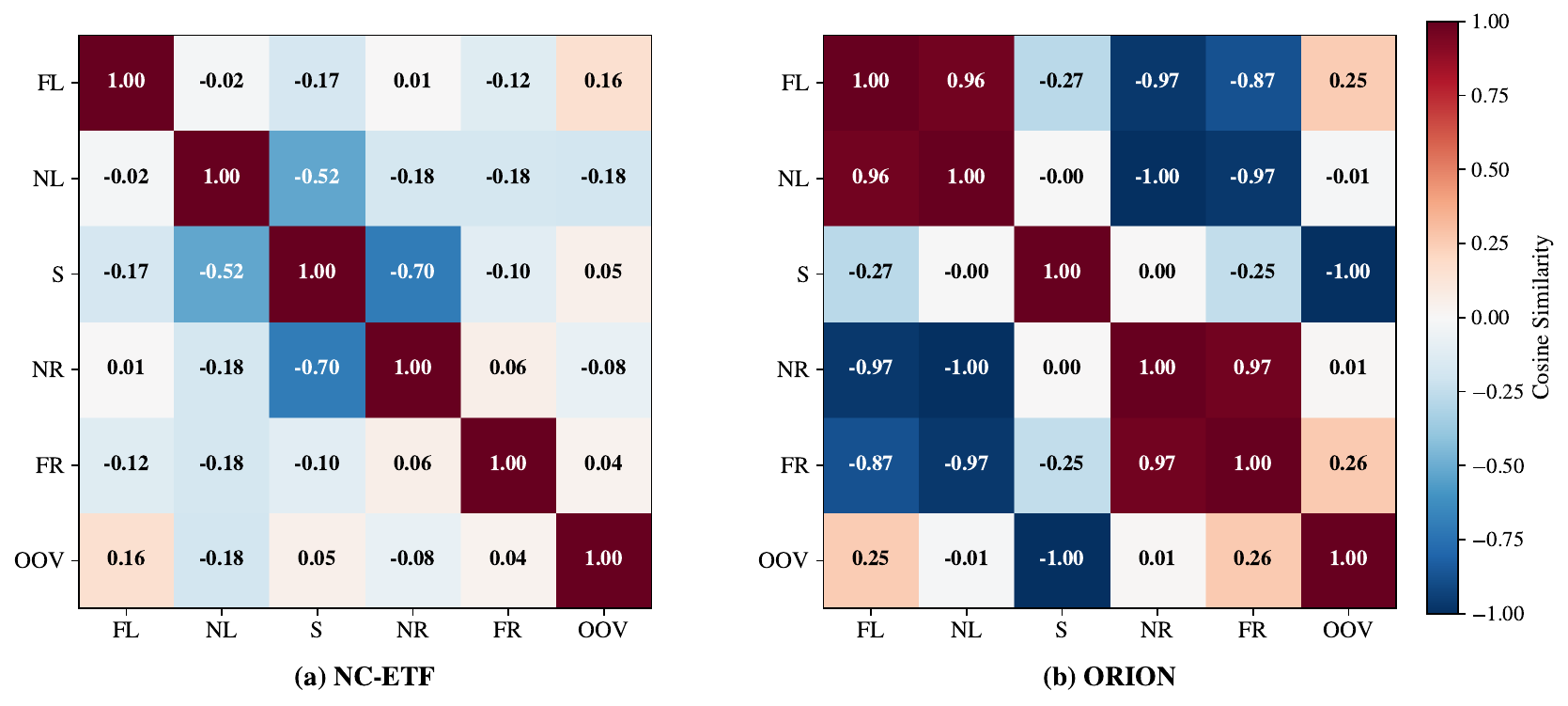}
    \caption{\textbf{Pairwise cosine similarities between class means after pretraining.}
    FL=Far Left, NL=Near Left, S=Straight, NR=Near Right, FR=Far Right, OOV=Out-of-View.
    (a)~NC-ETF fails to capture the natural relationships between navigation actions, treating classes as unrelated.
    (b)~ORION: exhibits a clear ordinal structure: adjacent classes are similar, while opposing classes are anti-correlated.}
\label{fig:fig_etf_mismatch}
\vspace{-0.5cm}
\end{figure}

\subsection{Class Mean Similarity Analysis}
\label{sec:etf_mismatch}

To examine how control classes are structured in the learned feature space, we compute pairwise cosine similarities between class mean features, producing the similarity matrix shown in Fig.~\ref{fig:fig_etf_mismatch}.
The Simplex ETF enforces a uniform pairwise cosine similarity of $-1/(C{-}1) = -0.2$ across all $C{=}6$ classes.
NC-ETF fails to reach this target, as pairwise cosines range from $-0.70$ to $+0.16$ with no clear relationship between semantic class distance and geometric similarity.
For example, \textit{Far Left}--\textit{Straight} ($-0.17$) and \textit{Far Left}--\textit{Far Right} ($-0.12$) yield nearly identical values, despite encoding very different navigational relationships.
ORION, by contrast, produces a monotonic ordinal structure where adjacent classes are similar and opposing classes are anti-correlated, consistent with the theoretical geometry for uniformly spaced class means along a one-dimensional axis.

An interesting pattern emerges for the Out-of-View (\textit{OOV}) class, which is excluded from the ordinal loss.
Under ORION, \textit{OOV} shows strong negative similarity with \textit{Straight} ($-1.00$) while being positively similar to extreme turning classes.
This behavior aligns with the navigation context, as the goal falls outside the field of view primarily when the robot needs to execute a sharp turn. Further qualitative analysis of the \textit{OOV} class is provided in Sec.~\ref{sec:linear_probe}.

\subsection{Linear Probe Analysis of Control Classes}
\label{sec:linear_probe}

\begin{figure}[t]
    \centering
    \includegraphics[width=\columnwidth]{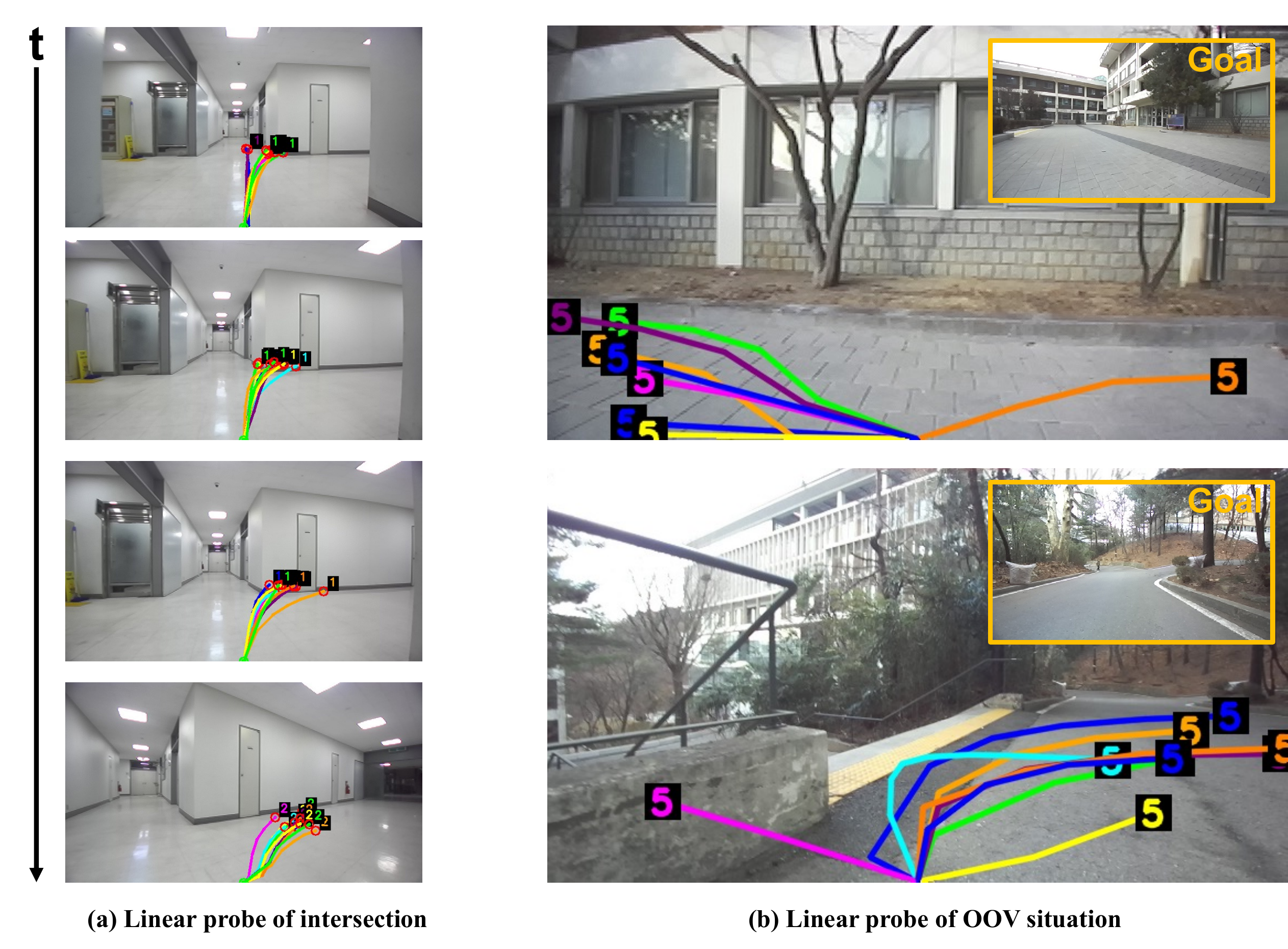}
    \caption{\textbf{Linear probe analysis of ORION.} Predicted trajectories and linear probe outputs are overlaid on the current observation. Classes 0--4 correspond to directions from \textit{Far Right} to \textit{Far Left}, and class 5 denotes OOV. (a) Intersection scenario. (b) OOV scenario.}
    \label{fig:linear_probe}
\end{figure}

To analyze how the learned representation relates to control decisions, we attach a linear probing head to the encoder features.
The probe predicts the control class from the encoder features.
This linear probe is used only for analysis and does not influence policy training.
We visualize the predicted control classes together with the generated trajectories in Fig.~\ref{fig:linear_probe}.
Classes 0--4 correspond to \textit{Far Right} to \textit{Far Left}, while class 5 denotes the \textit{OOV} case. 
Fig.~\ref{fig:linear_probe}(a) shows a sequence of inference results at an intersection. 
Consistent with the linear probe predicting right-turn classes (1 and 2), the generated trajectories repeatedly follow right-turn paths.
Fig.~\ref{fig:linear_probe}(b) illustrates an \textit{OOV} scenario where the goal image has little visual relationship with the current observation. 
In this case, the linear probe predicts class 5, and the generated trajectories correspondingly produce sharp turning motions.
\section{Implementation Details}
\label{sec:impl}

\subsection{Stable Statistics Estimation and Gradient Flow}
\label{sec:welford}

The ORION pretraining objectives depend on class means $\mu_k$ and within-class variances,
which involve \emph{ratios} of second-order statistics (Eq.~\ref{eq:welfords}) and axis estimation from $K$ class centroids.
These quantities are sensitive to sampling noise in individual mini-batches.
A small perturbation in a single class mean can cause large fluctuations in CDNV denominators 
and abrupt rotations of the ordinal axis $w$.
We address this with a dual-track design: \emph{running statistics} for stable estimation, 
and \emph{current-batch computation} for gradient flow.

\paragraph{Running statistics via Welford's algorithm.}
We maintain per-class running means $\bar{\mu}_k$, counts $n_k$, and sums of squared deviations $S_k$ using Welford's online algorithm~\cite{welford1962note}.
These are updated incrementally at each batch without storing historical features:
\begin{align}
    \label{eq:welfords}
    n_k &\leftarrow n_k + 1, \quad
    \delta = h - \bar{\mu}_k, \quad
    \bar{\mu}_k \leftarrow \bar{\mu}_k + \delta / n_k, \quad
    S_k \leftarrow S_k + \delta \cdot (h - \bar{\mu}_k).
\end{align}
These updates are performed with \texttt{torch.no\_grad()} and serve as stable reference values.
They are \emph{not} part of the computation graph.

\paragraph{Gradient flow through current-batch features.}
All loss terms (Eqs.~1, 3, 4 in the main paper) are computed from current-batch features $h_t = f_\theta(o_t)$, which retain their computation graph. The running statistics $\bar{\mu}_k$ and 
$\bar{\mu}$ enter only as \emph{detached constants}: wherever a class mean or global mean appears in the loss computation, the running estimate is used with stopped gradients. Gradients thus 
flow exclusively through $h_t$, updating the encoder parameters $\theta$ while the reference statistics remain stable across batches.

\paragraph{Axis smoothing.}
Additionally, the supervised ordinal axis $w$ is smoothed via exponential moving average (EMA, $\beta{=}0.95$) to prevent abrupt rotations from noisy class mean estimates.
The axis is detached from the computation graph; gradients for $\mathcal{L}_\mathrm{orth}$ flow through feature residuals, not through $w$ itself.

\subsection{Training Dynamics}
\label{sec:train_dynamics}

\begin{figure}[h]
 \includegraphics[width=\linewidth]{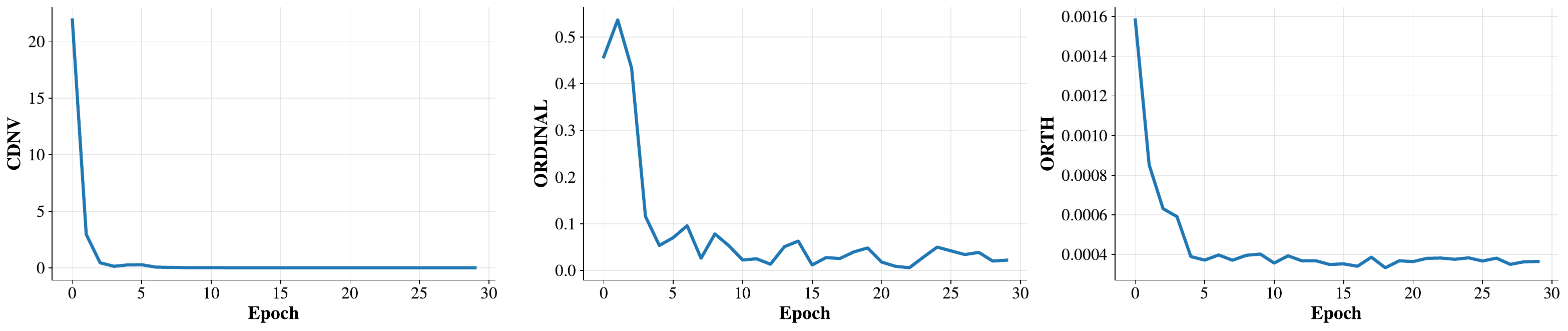}
\caption{\textbf{Stage~1 pretraining dynamics.}}
\label{fig:loss_orion}
\vspace{-0.8cm}
\end{figure}

Fig.~\ref{fig:loss_orion} shows individual loss components during ORION's Stage~1 pretraining. 

\subsection{Additional Qualitative Results}
\label{subsec:qual}

\begin{figure}[t]
    \centering
    \includegraphics[width=\columnwidth]{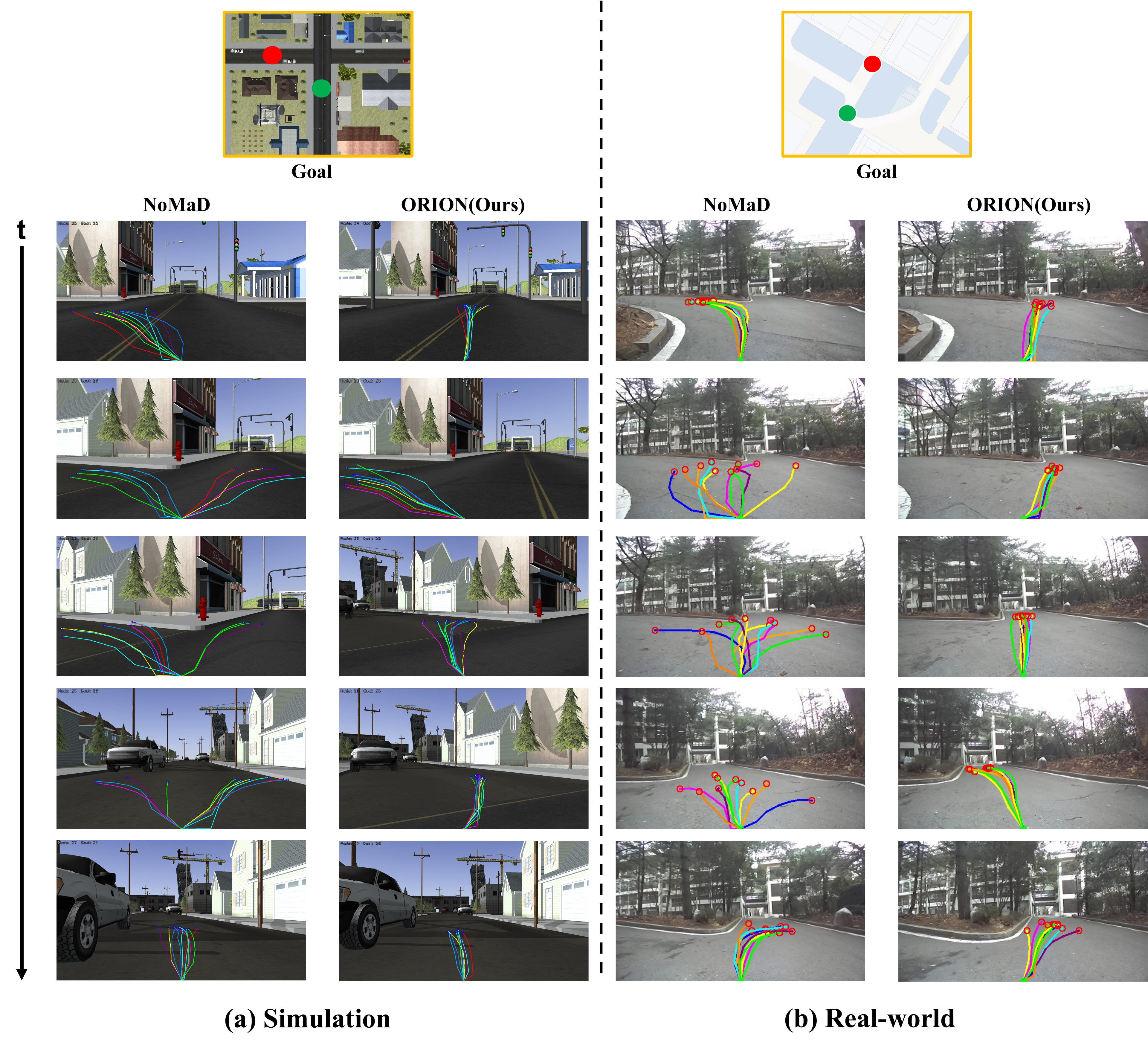}
    \caption{\textbf{Qualitative comparison of sequential inference results at an intersection.} (a) Simulation. (b) Real-world. The green dot indicates the robot's current position, while the red dot indicates the location of the given goal image.}
    \label{fig:seq_qual}
\end{figure}

\begin{figure}[H]
    \centering
    \includegraphics[width=\columnwidth]{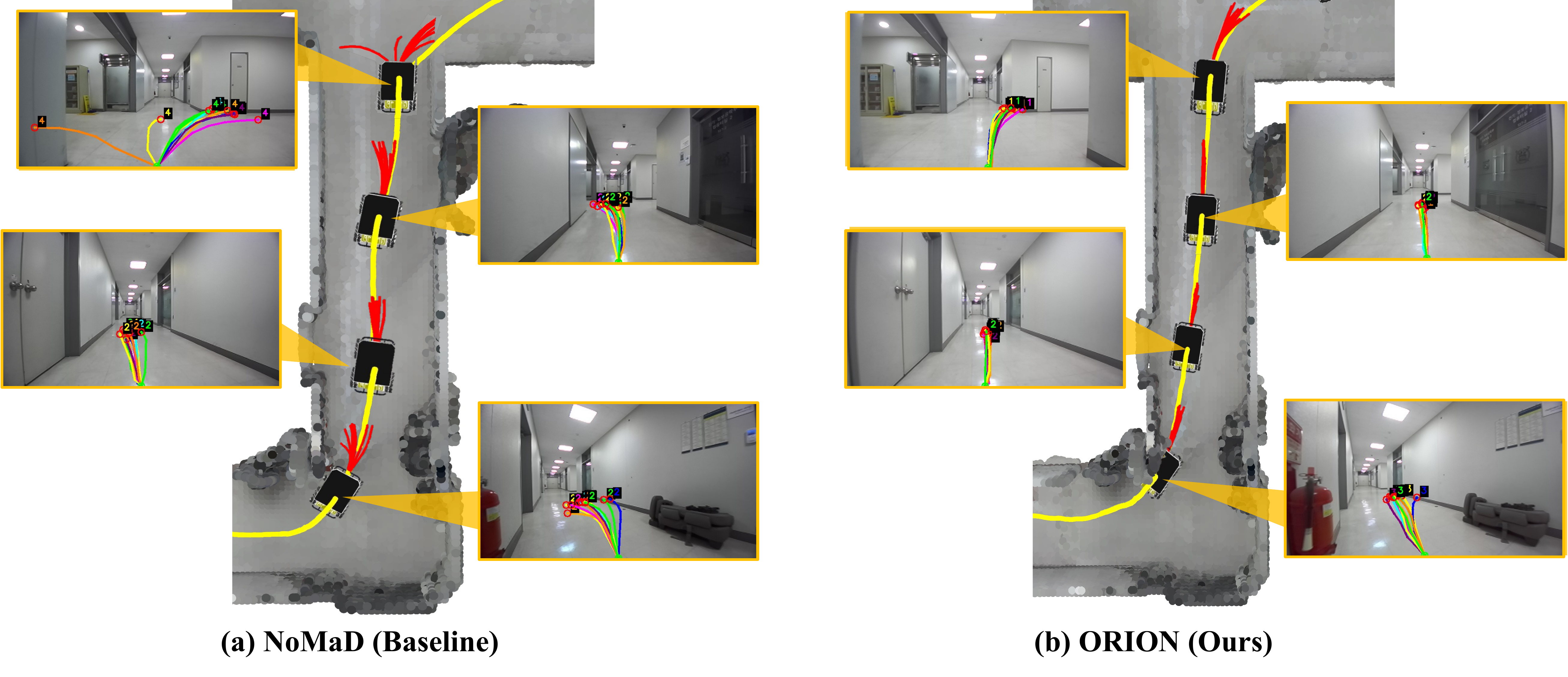}
    \caption{\textbf{Qualitative comparison along a real-world trajectory in a narrow corridor.} (a) NoMaD. (b) ORION. The yellow and red lines denote the robot trajectory and sampled actions, respectively.}
    \label{fig:seq_qual_indoor}
\end{figure}

We provide additional qualitative analysis in Fig.~\ref{fig:seq_qual} and Fig.~\ref{fig:seq_qual_indoor}. 
Sequential inference results show temporal consistency and stable goal-directed alignment in challenging intersections and narrow corridors.

\subsection{Deployment Setup}

\begin{figure}[!h]
    \centering
    \includegraphics[width=0.7\textwidth]{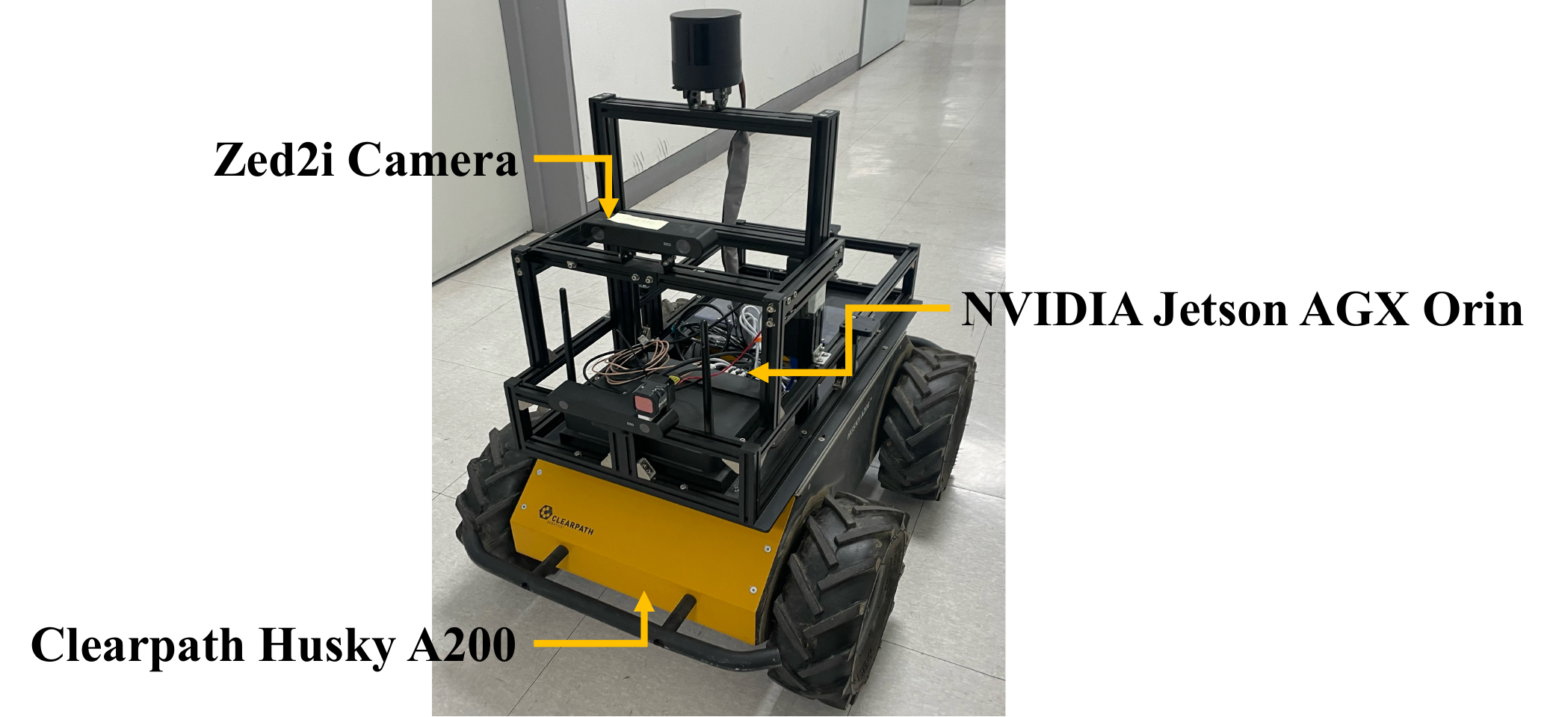}
    \caption{\textbf{Real-world robot platform used in our experiments.} The system uses an NVIDIA Jetson AGX Orin and a ZED~2i stereo camera (left RGB only), running the navigation model onboard at 3\,Hz with an average speed of 0.5\,m/s. Low-level control is performed by a PID controller with parameters consistently tuned across all baselines and environments.}
    \label{fig:husky}
\end{figure}

The robot platform used in our experiments is shown in Fig.~\ref{fig:husky}. 

%
%
\bibliographystyle{splncs04}
\bibliography{supplementary_refs}